# Nature Inspired Dimensional Reduction Technique for Fast and Invariant Visual Feature Extraction

**Ravimal Bandara[1], Lochandaka Ranathunga [2], Nor Aniza Abdullah [3]**
[1] University of Moratuwa, Sri Lanka., ravimalb@uom.lk
[2] University of Moratuwa, Sri Lanka., lochandaka@uom.lk
[3] University of Malaya, Malaysia, noraniza@um.edu.my

# Nature Inspired Dimensional Reduction Technique for Fast and Invariant Visual Feature Extraction

**Ravimal Bandara[1], Lochandaka Ranathunga [2], Nor Aniza Abdullah [3]**
[1] University of Moratuwa, Sri Lanka., ravimalb@uom.lk
[2] University of Moratuwa, Sri Lanka., lochandaka@uom.lk
[3] University of Malaya, Malaysia, noraniza@um.edu.my

**ABSTRACT**

Fast and invariant feature extraction is crucial in certain computer vision applications where the computation time is constrained in both training and testing phases of the classifier. In this paper, we propose a nature-inspired dimensionality reduction technique for fast and invariant visual feature extraction. The human brain can exchange the spatial and spectral resolution to reconstruct missing colors in visual perception. The phenomenon is widely used in the printing industry to reduce the number of colors used to print, through a technique, called color dithering. In this work, we adopt a fast error-diffusion color dithering algorithm to reduce the spectral resolution and extract salient features by employing novel Hessian matrix analysis technique, which is then described by a spatial-chromatic histogram. The computation time, descriptor dimensionality and classification performance of the proposed feature are assessed under drastic variances in orientation, viewing angle and illumination of objects comparing with several different state-of-the-art handcrafted and deep-learned features. Extensive experiments on two publicly available object datasets, coil-100 and ALOI carried on both a desktop PC and a Raspberry Pi device show multiple advantages of using the proposed approach, such as the lower computation time, high robustness, and comparable classification accuracy under weakly supervised environment. Further, it showed the capability of operating solely inside a conventional SoC device utilizing a small fraction of the available hardware resources.

## 1. INTRODUCTION

The recent advancement of machine learning has enabled learning of optimal features for applications by observing a large sample of images from the application domain. However, the feature learning and inference in modern approaches require an enormous amount of computer resources and time[1, 2] though visual computing on resource-constrained devices are trendy[3]. In addition to the computational cost, the state-of-the-art machine learned features are less invariant to dramatic geometrical transformation and illumination. Hence, generic handcrafted features can be more suitable for some applications which require invariant performance, faster execution with better adaptability [4, 5].

Visual feature description is one of the important stages in any visual understanding applications. It can be categorized into local and global feature descriptors[6, 7]. Throughout the past research works, several significant local descriptors have been invented for object recognition, such as Scale Invariant Feature Transform (SIFT) and its variants[8, 9], binary feature descriptors[10–12], and several global feature descriptors such as compacted dither pattern codes (CDPC)[13] and Gabor Pyramidal Histogram of Oriented Gradients (GPHOG)[14]. The high dimensionality of these descriptors negatively affects not only the classification performance but also the computation time and storage. This problem is well known as the "curse of dimensionality"[15]. Principal Component Analysis (PCA)[16], Locality Preserving Projections (LPP)[17] and fuzzy lattices technique[18] are frequently used dimensionality reduction techniques. PCA is widely used for dimensionality reduction of colors[19], descriptors such as SIFT (PCA-SIFT)[20] and embedded to some feature extraction algorithms like in GPHOG.

Conventional dimensionality reduction techniques require well-corresponded feature descriptions between the different visual contents to seek a meaningful low dimensional subspace[21]. Embedding these techniques in addition to the major feature extraction algorithm increases the computational time[20]. In this work, we re-introduce an existing color representation technique called dithering as a dimensionality reduction technique to efficiently extract and describe salient visual features.

Dithering is a popular image representation technique, which can be used to greatly reduce the number of colors in an image without affecting the overall appearance. Therefore,

This work was supported by National Research Council, Sri Lanka (Grant No. 12-017).

the technique has been mostly used in printers for color reproduction. The ability to exchange the spectral and spatial resolution by the human brain enables reconstructing the missing color information in dithered images. Conventional dimensional reduction techniques seek for a meaningful common subspace of a given set of data[21] whereas dithering reduces the spectral resolution with preserving the necessary spectral information without knowledge of data[22]. Therefore, in this work, we argue that dithering can be used as a fast and generic dimensionality reduction technique, which does not depend on the distribution of the data but still can be used to extract image features by utilizing a significantly smaller fraction of available hardware recourses. We employ this unconventional dimensionality reduction technique inspired by the human visual system, to construct a feature descriptor by simultaneously addressing the three main concerns i.e. the discriminative power, robustness, and the efficiency for real-time object recognition applications on most platforms including the resource-constrained devices.

Several beneficial properties of the concept which is behind the proposing technique were already studied in our previous works [23–25]. However, the performance related to object recognition has not been studied. In this study, we introduce a novel technique to use this dimensionality reduction concept for efficient object recognition. We use an extremely fast and low artifact error-diffusion technique[26] improved by an indexed searching mechanism for fast color quantization to form a set of color patterns. Then a set of salient feature points are extracted by a novel Hessian matrix-based analysis. The salient features are then described by a novel oriented spatial-chromatic descriptor. The proposed descriptor was evaluated for classification performance, invariance, and computational cost using two openly available object datasets namely coil-100[27] and ALOI[28] comparing with several state-of-the-art visual descriptors. The experiments were done on two hardware platforms, a conventional desktop computer without using any hardware acceleration and a single board SoC device which consists of small random-access memory.

In the next section, the strengths and weaknesses of the related state-of-the-art object recognition techniques will be discussed. The proposed technique is described in the third section and the remaining sections are allocated for describing the experimental setup, the results, discussion, and the conclusion.

## 2. RELATED WORK

Object recognition follows approaches from different categories such as patch based keypoint descriptor matching, global feature descriptor classification and deep learning-based classification. SIFT[8] and Speeded Up Robust Feature (SURF)[29] are the most used key point matching based techniques for object recognition. SIFT is scale and rotational invariant but suffers from a high computation time and dimensionality[16]. SURF lacks of discriminative power and the invariant properties exist with SIFT[16]. Both SIFT and SURF cannot be computed in real-time[16]. Binary descriptors such as Binary Robust Independent Elementary Features (BRIEF) and Binary Robust Invariant Scalable Keypoints (BRISK) are efficient [10, 11] but do not outperform the discriminative power of SIFT or SURF[12].

In order to obtain a global descriptor out of the set of unordered local features of objects, bag of feature (BoF) models are widely used[30–32]. BoF-SIFT, BoF-SURF and their color variants are well-known examples of highly discriminative BoF descriptors[33, 34]. Therefore, we use the BoF of SIFT and SURF and their variants in opponent color space to compare the performance of the proposed approach. The Pyramidal HOG (PHOG)[35] and Fused color Gabor-PHOG (FC-GPHOG)[14] are the improved versions HOG[36] which are frequently used global descriptors. FC-GPHOG is the state-of-the-art handcrafted features currently available[14]. Therefore, we evaluated the proposed descriptor comparing with FC-GPHOG as well.

Convolution Neural Network (CNN) is the state-of-the-art by means of the accuracy in visual data classification, out of all existing approaches [37]. CNN is a known high computer resource intensive approach; hence it typically can be found only in high-performance or accelerated hardware platforms. In this study, we compare our approach with a CNN model specified in [37] on a conventional desktop PC and a SoC device, without any hardware acceleration.

## 3. LOW DIMENSIONAL FEATURE-BASED OBJECT DESCRIPTOR

The core of this work is utilizing a color dithering as a dimensionality reduction technique in order to extract an object descriptor. The concept behind the dithering and the previous usage in feature extraction are explained in the next subsection.

### 3.1 Dithering for Feature Extraction and Description

The lower sensitivity of the human visual system to spatial resolution and the property of exchanging higher color resolution with lower spatial resolution have been used to overcome the incapability of reproducing true colors in printed media and some display devices [38]. Compacted Dither Pattern Code (CDPC)[13, 39] is the only study found in the literature related to dithering based visual data description, though it does not employ any dithering technique. The lack of spatial information results in limiting the usage of CDPC only for scene classification rather than object classification[13, 39]. We introduce the object classification capability to the proposed approach by embedding both chromatic and spatial details of salient dither patterns to the descriptor.

### 3.2 Salient Dither Pattern Feature Extraction

Color dithering introduces micro color patterns, which consist of a few different colors. These dither patterns create the illusion of the existence of many other colors, which do not exist in the actual image. The dither color patterns are different from their neighbor patterns at salient regions of an image. We could observe that these different patterns are reoccurring regardless of the orientation variations and slight scale variations. In [23–25], we have used a simple algorithm to prove the potentials of the concept. However, the algorithm was not competent enough to recognize an object with reasonable accuracy. Therefore, in this study, we introduce a completely novel technique to extract SDPF without sacrificing the gained benefits which have been described in [23–25].

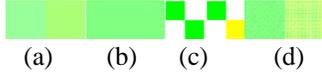

(a)     (b)     (c)     (d)

**Figure 1:** Colour contrast preserving the ability of dithering (a) two nearly similar colors with the values left: RGB(153,255,153) and right: RGB(171,255,119), (b) after applying linear color quantization to 12 color levels considering the hue component (c) the two color patterns created by the ED dithering algorithm (d) two regions with dithered colors preserving the color contrast.

The novel SDPF extraction algorithm consists of four major steps namely color quantization, calculating Hessian response, analyze the Hessian response to detect salient dither patterns and suppress the non-maximal. The quantization is done by using a fast error diffusion (ED) dithering technique explained in [26]. The ED-dithering technique creates different color patterns for near similar colors as shown in Figure 1, which ultimately preserves even the lower color contrast by using different color patterns.

The dithering algorithm has been implemented by improving the Floyd-Steinberg E-D algorithm. It has been identified a set of optimal error diffusion coefficients which depend on the input signal. The optimality has been defined where the Fourier spectrum as close as possible to the blue noise, hence it creates fewer artifacts. The computation is faster than the conventional Floyd-Steinberg E-D algorithm due to the smaller number of arithmetic operations and memory accesses. The dithering process includes two main sections namely the color quantization and error diffusion. The color quantization is done using (1).

$$N'_{00} = C_i \; ; \; C_i \in C_d, C_i \sim N_{00} \quad (1)$$

Where $N_{00}$ is the color of a pixel, $C_d$ is the dither color space and $N'_{00}$ is the new color of the pixel. The "~" symbol expresses that the two colors are the nearest in RGB space. The error should be calculated after the quantization using (2).

$$E = N_{00} - N'_{00} \quad (2)$$

Where E is the quantization error. Finally, the error is diffused to a set of neighbors by weighting as shown in (3).

$$N_{10} = D_{10}(N_{00}) \times E$$
$$N_{01} = D_{01}(N_{00}) \times E \quad (3)$$
$$N_{-11} = D_{-11}(N_{00}) \times E$$

Where $N_{10}$, $N_{01}$, and $N_{-11}$ are the right, bottom and bottom left neighbors and $D_{10}$, $D_{01}$ and $D_{-11}$ are the respective optimized error distribution coefficient vectors which can be found in [26].

In this study, the dither colors have been selected from the eight outermost corners of the RGB color cube as shown in Figure 2. The effect of the number of dither colors is discussed in section 3. This specific color selection brings the ability to index the colors for fast searching the closest color to the reference pixel value with only 3 logical comparisons using the binary search tree shown in Figure 3. The numbers ranging from 1 to 8 in Figure 3 are the indices used for representing the individual dither colors and (R,G,B) is the red, green and blue component of the pixels whereas the values of $R_h$, $G_h$ and $B_h$ are the centers of the range of R, G and B channels as shown in Figure 2.

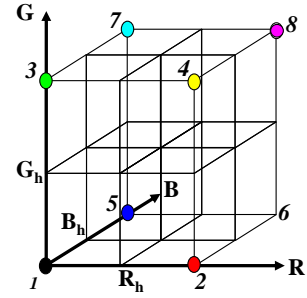

**Figure 2:** dither color set in RGB space

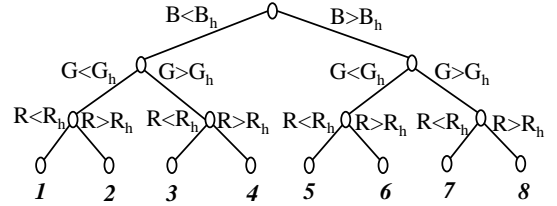

**Figure 3:** Binary search tree of dither colors

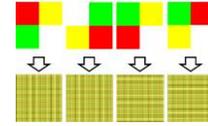

**Figure 4:** Color approximation of dither patterns with the same set of colors but with different color arrangements

The first level of error diffusion occurs among four adjacent pixels hence we consider the set of pixels as a dither pattern where each pattern contains four color values. We observed that the order of the colors within the pattern does not affect to the overall color approximation as shown in Figure 4. Therefore the color indices are sorted within each pattern to remove the permutations.

The structure of a pattern is shown in Figure 5. Any pattern, which has a great dissimilarity over its neighbor patterns, is defined as an SDPF. The dissimilarity of two dither patterns is specified as in (4).

$$d(P_{i,j}, P_{k,l}) = \sum_{c=0}^{3} w_c \quad (4)$$

$$w_c = \begin{cases} 1 & P_{ij}(c) \neq P_{k,l}(c) \\ 0 & otherwise \end{cases} \quad (5)$$

Where $d(P_{i,j}, P_{k,l})$ is the dissimilarity measurement of two patterns $P_{i,j}$ and $P_{k,l}$, and $P_{i,j}(c)$ is the $c^{th}$ color index out of four colors in the pattern $P_{i,j}$ and so on.

It has been proven that the critical points in a distribution can be obtained by analyzing the Hessian matrix [29]. Therefore, we introduce a technique, which characterizes a conventional Hessian matrix analysis approach with a novel approximation to obtain the second order derivative of a 2-dimensional color pattern distribution. The critical points, such that the local extrema of the dither pattern distribution are calculated by using the determinant of a Hessian matrix which is expressed by (6).

$$H(i,j) = \begin{bmatrix} L_{xx}(i,j) & L_{xy}(i,j) \\ L_{xy}(i,j) & L_{yy}(i,j) \end{bmatrix} \quad (6)$$

$$L_{xx}(i,j) = d(P_{i-1,j}, P_{i,j}) + d(P_{i,j}, P_{i+1,j})$$

$$L_{yy}(i,j) = d(P_{i,j-1}, P_{i,j}) + d(P_{i,j}, P_{i,j+1})$$

$$L_{xy}(i,j) = \frac{1}{4}\big(d(P_{i-1,j-1}, P_{i-1,j+1}) + d(P_{i+1,j-1}, P_{i+1,j+1}) + d(P_{i-1,j-1}, P_{i+1,j+1}) + d(P_{i-1,j+1}, P_{i+1,j+1})\big)$$

The determinant is calculated using (7).

$$det(H(i,j)) = L_{xx}(i,j) \times L_{yy}(i,j) - L_{xy}(i,j)^2 \quad (7)$$

The potential SDPF points are selected by applying a threshold for the determinant of the Hessians of patterns in order to filter out the most stable extremes in the distribution. The potential SDPF points are selected by (8).

$$D(i,j) > T(i,j) \rightarrow \big(P_{i,j}, D(i,j)\big) \in S_s \quad (8)$$

$$D(i,j) = abs\big(det(H(i,j))\big) \quad (9)$$

Where $S_S$ is the set of potential feature points and $T(j,j)$ is the threshold obtained using (10).

$$T(i,j) = d(P_{i-1,j-1}, P_{i,j-1}) + d(P_{i,j-1}, P_{i+1,j-1}) + d(P_{i+1,j-1}, P_{i+1,j}) + d(P_{i+1,j}, P_{i+1,j+1}) + d(P_{i+1,j+1}, P_{i,j+1}) + d(P_{i,j+1}, P_{i-1,j+1}) + d(P_{i-1,j+1}, P_{i-1,j}) + d(P_{i-1,j}, P_{i-1,j-1}) \quad (10)$$

The equation (10) was obtained empirically as it yields better classification accuracy.

The non-maximal patterns then suppressed by using (11),

$$\forall \big(P_{i,j}, D(i,j)\big) \in S_s$$
$$\big(R_p \subset S_s \text{ AND } \forall \big(P_{i',j'}, D(i',j')\big) \in R_p : D(i,j) \geq D(i',j')\big) \rightarrow \big(P_{i,j}, D(i,j)\big) \in S_f\big) \quad (11)$$

where $R_p$ be the set of potential feature points within the window centered at the pattern $P_{i,j}$, $(P_{i,j}, D(i,j))$ and $(P_{i',j'}, D(I',j'))$ are the ordered pairs of potential feature points and their strengths respectively, and $S_f$ be the set of SPDF. The extracted SDPF is described by a 3D spatial-chromatic histogram which contains not only the spatial and color information but also their correlation. The $R_p$ is a square shaped window which empirically set to 5 throughout all the experiments.

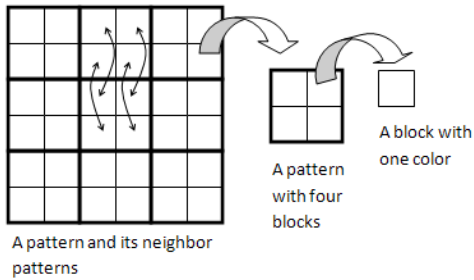

**Figure 5:** Structure of the dither pattern: Any of four adjacent color blocks is considered as a pattern

### 3.3 Spatial-Chromatic Histogram of SDPF

The chromatic information is obtained by using the dither colors in SPDF where the spatial information is obtained using the centroid distances and the angles which are made by each SDPF points relative to the dominant orientation.

#### 3.3.1. Centroid Distance Measurement

The centroid distance function is a known shape representation technique[40]. It is invariant to the rotation and translation yet resists to noise and occlusions [41]. The centroid is calculated for the set of SDPF in $S_f$ using (12).

$$x_c = \frac{\sum_{n=1}^{N} x(n)}{N}, y_c = \frac{\sum_{n=1}^{N} y(n)}{N} \quad (12)$$

Where $(x_c, y_c)$ is the centroid, $x(i)$ and $y(i)$ are the x and y coordinates of $n^{th}$ point in $S_f$, $N=|S_f|$.

The centroid distance function r(n) is expressed by the distance of the SDPF points from the centroid $(x_c, y_c)$ of the shape as in (13).

$$r(n) = ([x(n) - x_c]^2 + [y(n) - y_c]^2)^{\frac{1}{2}} \quad (13)$$

The encoding of the spatial details of SDPF is done by preparing the distance axis of the histogram. Let $D_f^2(n)$ is the squared value of the distance from $n^{th}$ feature point to the centroid and $\max\big(D_f^2(n)\big)$ denotes the maximum of $D_f^2(n)$ value; $k_d$ is the number of bins going to be in the histogram. The upper boundary of the range of the $i^{th}$ bin $R_d(i)$ is defined as in (14) with normalizing the centroid distances of SDPF points.

$$R_d(i) = \begin{cases} \frac{\max\big(D_f^2(n)\big)}{k_d} & i = 1 \\ R_d(i-1) + \frac{\max\big(D_f^2(n)\big)}{k_d}, & 1 < i \leq k_d \end{cases} \quad (14)$$

The calculation of bin ranges requires only the squared of centroid distance $(D_f^2(i))$ as shown in (14) and it is equal to the squared of function r(n) hence we use (15).

$$D_f^2(n) = [x(n) - x_c]^2 + [y(n) - y_c]^2 \quad (15)$$

This adoption omits a large number of square root operations to save computational time. After preparing the bins, each SDPF in $S_f$ is allocated to one of the k distance bins. The distance bin $B_d$ of each SDPF point can be obtained using (16).

$$B_d = i \; ; \; R_d(i-1) < D_f^2(n) < R_d(i) \quad (16)$$

### 3.3.2. Orientation Normalization

The angles of SDPF points are also taken into consideration while constructing the histogram in order to increase the discriminative power of the final descriptor. An orientation normalization is required for making it rotational invariant. Therefore, the angles are calculated relative to the dominant orientation of the SDPF point distribution. The dominant orientation is calculated by using least square method (17).

$$m = \frac{\sum_{n=1}^{N}(x(n)-x_c)(y(n)-y_c)}{\sum_{n=1}^{N}(x(n)-x_c)^2} \quad (17)$$

Where $m$ is the slope of the best-fitted line. However, there is another ambiguity since the same $m$ can occur when the object turns upside down by 180 degrees, therefore a reference angle is calculated by considering the density difference of SDPF points in the two sides of the perpendicular line of the best-fitted line, which crosses through the centroid. The perpendicular line can be obtained by (18).

$$y = \frac{-1}{m}(x - x_c) + y_c \quad (18)$$

SDPF points can be classified into either of side by (19).

$$P_{side}(n) = \begin{cases} 0 & l(n) = 0 \\ 1 & l(n) > 0 \\ 2 & l(n) < 0 \end{cases} \quad (19)$$

$$l(n) = y(n) + \frac{1}{m}(x(n) - x_c) - y_c \quad (20)$$

Where $P_{side}(n)$ is an SDPF point and $x(n)$ and $y(n)$ are its coordinates. Then the starting angle can be obtained by (21).

$$\theta_0 = \begin{cases} \tan^{-1} m & |U_1| > |U_2| \\ \tan^{-1} m - 180^0 & |U_1| < |U_2| \end{cases} \quad (21)$$

Where $U_1$ and $U_2$ are the sets of SDPF points classified into the two sides and $||$ denotes the cardinality of a set. If $\theta_0$ is higher than 360 or less than 0, it will be revolved clockwise or counterclockwise by $360^0$ degrees respectively. The orientation normalized angle of an SDPF point is calculated by (22).

$$P_\theta(n) = \tan^{-1}\left(\frac{y(n)-y_c}{x(n)-x_c}\right) - \theta_0 \quad (22)$$

Where $P_\theta(n)$ is the angle of the $n^{th}$ SDPF point relative to the starting angle. If $P_\theta(n)$ is higher than 360 or less than 0, it will be revolved clockwise or counterclockwise by $360^0$ degrees respectively. Finally, the SDPF points are assigned to the angle bins ($B_a$) considering the normalized angles calculated for each of the SDPF points. The optimal amount of angle bins ($k_a$) were selected empirically, which will be discussed in the next section. The point allocation to angle bins is done as in (23).

$$B_a(n) = \left\lfloor \left(\tan^{-1}\left(\frac{y(n)-y_c}{x(n)-x_c}\right) - \theta_0\right)/R_a \right\rfloor \quad (23)$$

$$R_a = 360/k_a \quad (24)$$

Figure 6 shows a visualization of the instances of a sample object processed in different steps in the SDPF algorithm.

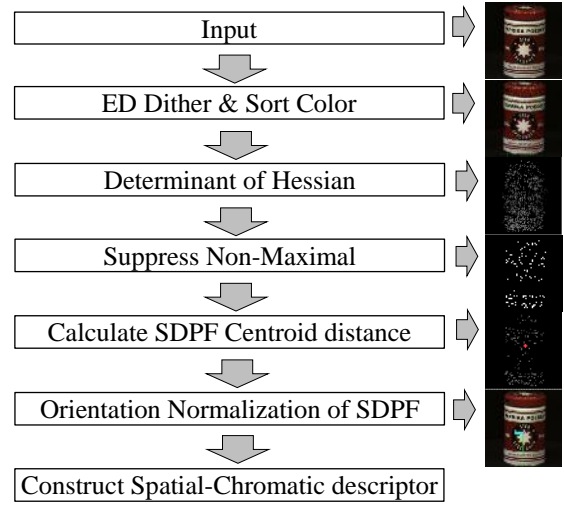

**Figure 6:** The SDPF process with the different instances of an object.

Once all necessary properties (i.e. colors in each of the salient dither patterns and normalized angle to the dominant orientation) of the extracted SDPF points are obtained the final descriptor can be constructed as explained in the next section.

### 3.3.3. Populating the SDPF Spatial-Chromatic histogram

Each SDPF point contains three properties such that the centroid distance, the angle relative to the dominant orientation and color pattern which contains four colors from the dither color space as shown in (25) and (26).

$$P_{(x,y)} = \{B_d, B_a, B_c\} \quad (25)$$

$$B_c = \{c_1, c_2, c_3, c_4\} \quad (26)$$

where $c_1...c_4$ are the four-color indices in each pattern counted in the order of top-left, top-right, bottom left and bottom-right cells.

The histogram contains color axis with the 8 color bins, centroid distance axis with 4 distance bins and 8 orientation bins. The centroid distance bins and the angle bins are obtained directly by $B_d$ and $B_a$ respectively. However, each of the patterns contributes four times each based on one of the four colors in the pattern, to a single or multiple color bins as shown in (26).

### 3.3.4. Dimensionality Optimization of the Histogram

The resultant histogram is used for object classification in images with support vector machine (SVM). An experiment was conducted to find the optimal value for the number of distance bins $k_d$, the number of angle bins $k_a$ and the number of color bins $k_c$. The average classification rate of images from Caltech dataset[42] was taken into consideration with all the combination of values of $k_d$ (ranges from 3 to 10), $k_a$ with the amounts of 4,8,12 and 15, and the number of color bins with the amounts of 6, 8, 14 and 26. We used 10 visual concept categories and selected 40% of images from each category to train SVM whereas the remaining 60% to obtain the classification rate.

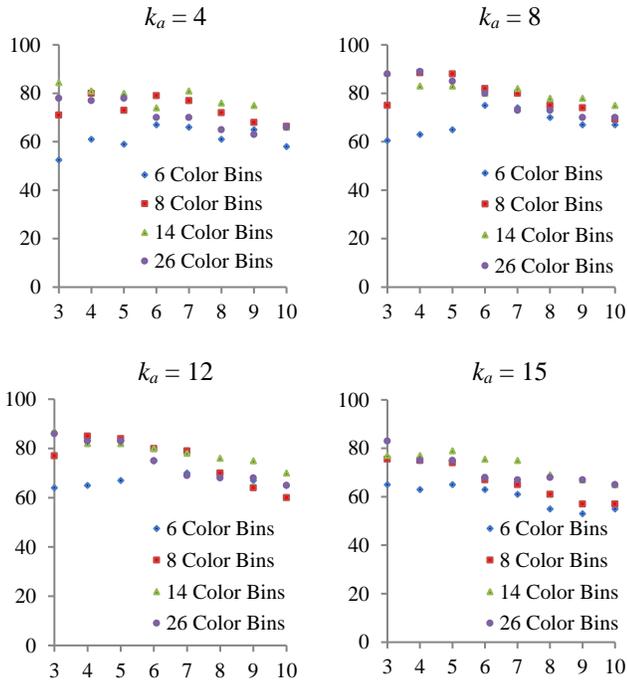

**Figure 7:** Average classification rate vs. number of distance bins and color bins

7 shows the average classification rates for each combination of different values of $k_d$, $k_a$, and $k_c$. The maximum accuracy found in the experiment is 89% where $k_d=4$, $k_a=8$ and $k_c=26$. The other option was 85.5% at both $k_d=3$, $k_a=8$, $k_c=14$ and $k_d=4$, $k_a=8$, $k_c=8$. Considering the lower dimensionality, the reasonable accuracy and the ability of fast computation, the instance where $k_d=4$, $k_a=8$ and $k_c=8$ was selected. All the experiments were conducted using this dimensional configuration. The results will be discussed in the next section.

### 3.4 Classification of SDPF Descriptor

We observed that SDPF descriptors are not linearly separable in its 256-dimensional Euclidean space, by testing the SDPF descriptors classifying using nearest neighbor classification. Therefore, we used a support vector machine with a polynomial kernel[43]. We include the training and classification performance of SDPF using C-SVM on the two specified platforms.

### 4. EXPERIMENT

We compared the performance of proposed descriptor with several feature point-based descriptors namely the bag of feature descriptions of SIFT (Bof-SIFT) and SURF (Bof-SURF), and their opponent color variants i.e. BoF-OSIFT, BoF-OSURF, FC-GPHOG and a convolution neural network (CNN). The BoF codebooks were obtained as described in [44]. We use codebooks with the size of 200 which has been used for SIFT and SURF features in several studies and yielded good results[32, 45]. We used the codebooks with the size of 600 for all color variant of BoF descriptors. The CNN was trained from the scratch by using the model specified in [37] which is the base of many successful deep neural network models used for large scale object and scene classification. The nearest neighbor rule is used as the classifier for FC-GPHOG whereas the SVM is used with the bag-of-feature descriptors.

Three separate experiments were carried to assess different aspects of the proposed descriptor. First, we measured the computational time of its feature extraction, training and classification processes on both desktop PC and Raspberry Pi. Secondly, we compared its classification performance using two datasets which we will describe in detail in the following section. Thirdly, we assessed the robustness of the descriptor in recognizing objects with variances in orientations of objects, viewing angles and illumination conditions.

### 4.1 Datasets

The experiments were conducted on two open datasets namely coil-100 data set and ALOI datasets. The coil-100 data set consists around 7200 different images of 100 man-made objects, each appears 72 times in different orientations ranging from 0 to 360 degrees around the object's vertical axis. The objects are well illuminated, background filled with a solid color and cropped around. We used coil-100 data set to evaluate the robustness of the SDPF descriptors to different orientations of the same object since the majority of the objects in coil-100 dataset is not symmetric along the orientation axis.

We used two image sets from ALOI dataset, one consists 24000 images captured under different illumination conditions (ALOI-ill) whereas the other set has 72000 images which have been captured under different view angles (ALOI-View). Both the sets have 1000 object categories. These two ALOI datasets allow us to evaluate the robustness of SDPF in recognizing objects under different illuminations and viewing angles. Further, these datasets can reveal the usability of our low dimensional SDPF descriptor to accurately recognize objects from a large number of object categories. The objects in either of the datasets do not have variations in the orientation around the axis towards the camera, hence the ALOI-View dataset has been augmented by including some randomly selected images rotated by 4 different degrees of angle.

### 4.2 Experiment Procedure

As the main objective of the study is to assess the suitability of the descriptor for the use in non-accelerated hardware platforms, the feature extraction and classification were done without using GPU, SIMD or multi-core acceleration. However, we utilized the GPU and SIMD registers in the training phase of CNN to reduce the time spent in the training. Both CNN and SVM were trained using a desktop PC which is equipped with a GTX770 CUDA GPU. The test was done using a desktop PC which is equipped with Intel i7 2600 3.4GHz, 8GB of RAM and with Ubuntu for CNN, and Windows 7 for SVM, and a Raspberry Pi 2 model B which is equipped with 900MHz quad-core ARM Cortex-A7 CPU, 1GB of RAM and Raspbian OS installed in a class 10 micro SD card. Note that we used Caffe deep learning framework to implement the CNN in Ubuntu, and an OpenCV based deep neural network implementation in Raspbian OS due to the technical difficulties arose while installing Caffe in Raspbian OS. The images were resized to 128x128 in order to reduce the complexity of the network model and memory consumption.

### 4.3 Evaluation Criteria

Average computation time for feature extraction was measured by using both coil-100 and ALOI datasets. The training and classification time was measured by using only the ALOI-View dataset which contains the highest number of images from the highest amount of object categories. The average extraction time and average classification time measured in milliseconds whereas the training time measured in minutes due to the values from the two measurements are having incomparable ranges. The usage of computer resources by the other programs and background services were kept consistent for all the experiments conducted for measuring the computational performance.

All the classifiers were trained using only 40% of images in each of the object categories and the remaining 60% were used to assess the classification performance. The classification performance was measured by the average precision using (27).

$$Average\ Precision\ (AP) = \frac{\sum_{n=1}^{N} \frac{|R_n \cap A_n|}{|A_n|}}{N} \times 100 \quad (27)$$

where $R_n$ is the set of relevant images to the n[th] object class, and $A_n$ is the resulting image set which is classified into the n[th] object class by the given classifier.

The third phase of the experiment was done in two ways, one with the original dataset and then with the augmented data set. In both cases, AP is calculated using (27).

### 4.4 Computational performance

Table 1 shows the average computation performances of the descriptors and their selected dimensionalities. The effect of the dimensionality of the descriptors to the computational time can be seen in this table. The significance of these results will be discussed in the next section.

Table 1. Dimension and average computational performance of the feature descriptors

| Method | Dimension | Desktop PC | | | RPi | |
|---|---|---|---|---|---|---|
| | | Extraction (ms) | Classification (ms) | Training (minutes) | Extraction (ms) | Classification (ms) |
| **SDPF** | 256 | **10** | 78.1 | **45** | **15** | 80.3 |
| BoF-SIFT | 200 | 470 | 77.2 | **45** | 710 | 79.8 |
| BoF-SURF | 200 | 380 | 77.2 | **45** | 530 | 79.8 |
| BoF-OSIFT | 600 | 1120 | 83.2 | 52 | 1610 | 90.7 |
| BoF-OSURF | 600 | 1110 | 83.2 | 52 | 1530 | 90.7 |
| FC-GPHOG | 12000 | 2590 | 7772 | 122 | 3820 | 10230 |
| CNN | 253440 | - | 1362 | 335 | - | 38200 |

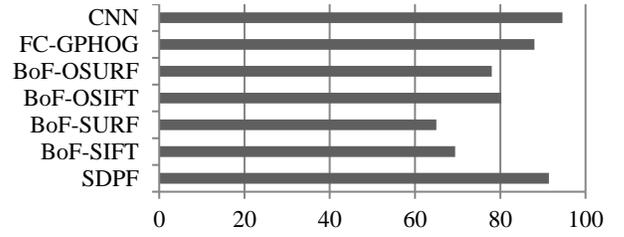

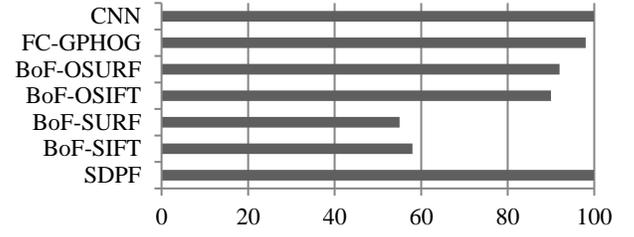

**Figure 8:** Average precision of classifying objects captured under random view angles; (a) ALOI data set, (b) Coil-100 dataset

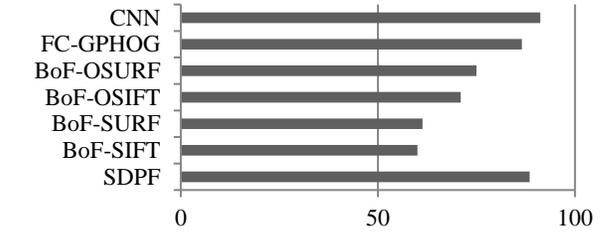

**Figure 9:** Average precision of classifying objects from ALOI-ill captured under random illumination conditions

### 4.5 Classification Performance

Figure 8 shows the average precision of recognizing objects in both ALOI-View and Coil-100 datasets where Figure 9 shows the precision of object recognition under different illumination with ALOI-Ill dataset.

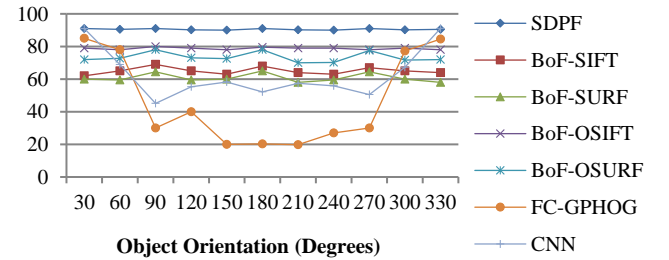

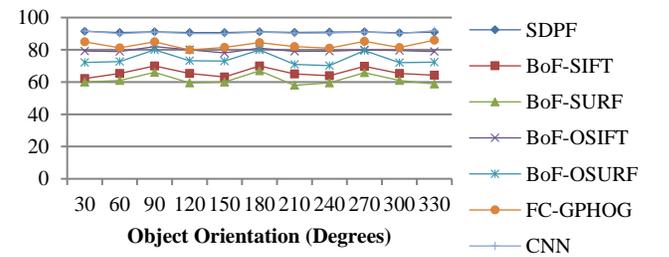

**Figure 10:** Average precision vs classifying objects with a different orientation. (a) Using a model which has been trained with non-augmented data from ALOI-View. (b) Using a model which has been trained with augmented data (0º, 90º, 180º, 270º) from ALOI-View.

## 4.6 Performance Under Weakly Supervised Environment

shows the average precision when an object has been rotated different angles with and without augmenting the dataset. This result was obtained to comparing the robustness of the proposed approach to perform under weakly supervised environments.

## 5. DISCUSSION

Table 1 shows that SDPF has achieved the best feature extraction time in both hardware platforms. The negligible gap of the extraction time of SDPF in between the PC and Raspberry Pi proves the better utilization of hardware resource by the SDPF algorithms. The result also expresses the proportional relationship between the dimensionality and the classification time, which explains the slightly higher classification time of the SDPF as well. However, the advantage of the lower dimensionality of all the bag of features-based descriptors does not improve their overall performance due to the extremely high feature extraction time. SDPF also achieved the second lowest dimensionality out of the all conventional descriptors while not scarifying the feature extraction efficiency. Although the dimensionality of the SDPF is slightly higher than the BoF-SIFT and BoF-SURF, the proposed descriptor includes both the color and spatial information. The fact that the training phase of CNN involves in optimizing 60 million parameters proves the excessive time taken for the training even utilizing the high-performance computational resources. The extremely long classification time of CNN prevents it using in RPi device for almost any application. Besides the classification time, the CNN takes another 20 seconds to load the network model at the initialization.

Table 2. The computational performance of SDPF with the breakdown of all major steps in the algorithm. ET is execution time. add: additions and substractions, mul: multiplications, div: divisions, sqrt: squareroot, cmp:logical comparison, tri: inverse trigonometric function.

| Description | Arithmetic operations per pixel/pattern | | | | | | ET (ms) |
|---|---|---|---|---|---|---|---|
| | add | mul | div | sqrt | cmp | tri | |
| ED-Dithering | 3 | 9 | 0 | 0 | 3 | 0 | 3.51 |
| Colour sorting | 0 | 0 | 0 | 0 | 6 | 0 | 0.35 |
| Calculate Hessian | 6 | 3 | 0 | 0 | 8 | 0 | 1.87 |
| Analyse Hessian | 7 | 0 | 0 | 0 | 10 | 0 | 1.50 |
| Non-max suppression | 0 | 0 | 0 | 0 | 8 | 0 | 0.23 |
| Centroid | 2 | 0 | 0 | 0 | 0 | 0 | 0.17 |
| Centroid distance | 3 | 2 | 0 | 0 | 0 | 0 | 0.22 |
| Distance bin ranges | 0 | 0 | 0 | 0 | 1 | 0 | 0.01 |
| Dominant orientation | 5 | 2 | 0 | 0 | 0 | 0 | 0.28 |
| Resolving upside down | 6 | 1 | 1 | 0 | 1 | 1 | 0.85 |
| SDPF angles | 3 | 0 | 1 | 0 | 1 | 1 | 0.71 |
| Descriptor construction | 0 | 0 | 1 | 0 | 4 | 0 | 0.21 |

The best overall computation time and the lower dimensionality together prove the superiority of the color dithering-based dimensionality reduction approach for object recognition from a large number of object categories, in resource-constrained devices. The algorithmic complexity and the details of atomic operations in the algorithm should be analyzed to measure the consistency of the execution on a typical hardware platform against the size of the input. Therefore, we have further broken down the computation time consumption to major steps in SDPF algorithm in order to identify the critical subprocesses.

Table 2 shows the computational details of every major step involved in the SDPF algorithm. The execution time is calculated by averaging the total time taken for obtaining SDPF descriptor of 100 images with the dimension of 128x128 on the desktop PC. Note that the majority of the operands of the arithmetic operations mentioned in the table are single-precision floating-point values. Only the ED-Dithering step operates for each and every pixel in the image. All the other steps, starting from the color sorting, up to the non-maximal suppression, operate on all the patterns. The number of patterns equals only to a quarter of the number of pixels. All the remaining steps operate on all the SDPF point which in amount is extremely less than all available patterns. The analysis shows that all the steps work in O(n) complexity and safely avoided the computationally expensive operations such as square root calculation. The linear complexity assures that the algorithm can comfortably handle larger images as well. The breakdown of the computation time reveals that the ED-dithering process consumes most of the time required to complete the whole process. Nonetheless, the required amount of arithmetic operations for the dithering process is comparable to other conventional ED-dithering approaches discussed in [26].

The graphs in Figure 8 show the average precision values exhibited by the seven methods when classifying ALOI-View and Coil-100 datasets. Both Figure 8 (a) and Figure 8 (b) clearly show that the proposed method outperforms all BoF based methods and FC-GPHOF while closely competing with the state-of-the-art precision of the CNN method. Although the proposed method is 3% behind the precision of the CNN for the ALOI-View, Figure 8 (b) shows that both methods achieved almost similar precision for Coil-100 dataset due to the absence of occluded objects.

Various illumination conditions can significantly affect the classification performance of any of the existing methods.

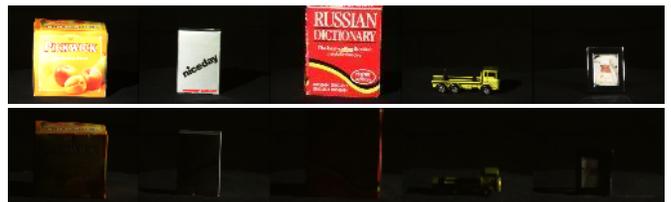

**Figure 11:** Sample images from five object categories in ALOI-ill dataset. Each column contains two images from a single object category, captured under different illumination conditions.

Therefore, we evaluated the proposed method under different illumination conditions and the results are shown in the graph in Figure 8. The graph shows that the proposed method has achieved an average precision closer to the CNN whereas outperformed the other baseline methods. Some of the objects are partially visible due to non-uniform illumination as shown in Figure 11. Many of the object categories contain several images with almost non-visible objects which cannot be distinguished even by a human. We believe this fact reduced the classification performance. The proposed method uses the centroid of an object hence theoretically the descriptor can be largely affected by the disappearance of any part of the object boundaries. However, the overall result shows that the classification performance of occluded objects, still comparable with the baseline methods. We believe that the color information embedded into the descriptor by using the color dithering technique in addition to the shape abstraction approach which is based on centroid distance and relative angle of SDPF in our method has contributed to this reliable performance with the occluded objects.

Figure 10 (a) shows that the SDPF has not significantly affected by the changes in objects' orientation whereas the FC-GPHOG and CNN show weaker classification performance. This implies the FC-GPHOG and the convolution features are less usable under weakly supervised training approach.

Figure 10 (b) denotes the results of the classifiers which were trained by using the ALOI-view data set augmented by the orientations of 0º, 90º, 180º and 270º. It shows that the augmentation significantly improved the classification performance of both FC-GPHOG and the CNN. The data augmentation significantly increases the size of the dataset which also results in longer training time. However, the overall result implies that SDPF outperforms all baseline methods in the weakly supervised environment. This property enables the SDPF based object recognizers to be trained quickly with a smaller number of object samples yet to perform reliably on many unseen instances of the given objects.

In summary, the experiments have shown that the proposed method outperforms the state-of-the-art methods by means of computational performance with less hardware resource requirement while achieving a closer classification performance to the state-of-the-art methods. The SPDF also poses superior robustness to unseen orientations of objects and drastic variances in illumination compared to all BoF based descriptors and FC-GPHOG. The results further elaborate that the SDPF is a practically feasible solution for object recognition in the applications where CNN based approaches cannot be used due to their requirement of the enormous amount of computer resources.

## 6. CONCLUSION

In this paper, we presented an approach to rapidly extract an invariant visual feature for object recognition by using a novel dimensionality reduction technique inspired by a natural phenomenon exists in the human visual system. The experimental results yield the conclusion, that the proposed method is advantageous where a highly invariant object recognition performance is required on resource-constrained devices.